\documentclass{article}

    \PassOptionsToPackage{numbers, compress}{natbib}

    \usepackage[final]{tackling_climate_workshop_style}

\usepackage[utf8]{inputenc} 
\usepackage[T1]{fontenc}    
\usepackage{hyperref}       
\usepackage{url}            
\usepackage{booktabs}       
\usepackage{amsfonts}       
\usepackage{nicefrac}       
\usepackage{microtype}      
\usepackage{graphicx}
\usepackage{wrapfig}
\usepackage{multirow}

\title{Discovering EV Charging Site Archetypes Through Few Shot Forecasting: The First U.S.-Wide Study}

\author{%
  Kshitij Nikhal \\
  Alpha Grid\\
  Palo Alto, CA 94306 \\
  \texttt{technikhal@alphagrid.ai} \\
  \And
  Lucas Ackerknecht \\
  Alpha Grid\\
  Palo Alto, CA 94306 \\
  \texttt{luke@alphagrid.ai} \\
  \AND
  Benjamin S. Riggan \\
  University of Nebraska-Lincoln\\
  Lincoln, NE, 68588 \\
  \texttt{briggan2@unl.edu} \\
  \And
  Phillip Stahlfeld \\
  Alpha Grid\\
  Palo Alto, CA 94306 \\
  \texttt{phil@alphagrid.ai} \\
}

\begin{document}

\maketitle

\begin{abstract}
 The decarbonization of transportation relies on the widespread adoption of electric vehicles (EVs), which requires an accurate understanding of charging behavior to ensure cost-effective, grid-resilient infrastructure. Existing work is constrained by small-scale datasets, simple proximity-based modeling of temporal dependencies, and weak generalization to sites with limited operational history. 
 To overcome these limitations, this work proposes a framework that integrates clustering with few-shot forecasting to uncover site archetypes using a novel large-scale dataset of charging demand. The results demonstrate that archetype-specific expert models outperform global baselines in forecasting demand at unseen sites. By establishing forecast performance as a basis for infrastructure segmentation, we generate actionable insights that enable operators to lower costs, optimize energy and pricing strategies, and support grid resilience critical to climate goals.
  
\end{abstract}

\section{Introduction}

The global transportation sector is undergoing a paradigm shift, with electric vehicles
(EVs) central to achieving net zero emissions in the United States~\cite{ahmadi2019environmental, woody2023decarbonization}. Achieving this transition requires not only rapid deployment of charging infrastructure, but also a nuanced understanding of EV charging behavior. Such insights directly inform investment decisions~\cite{linjuan2024site}, grid stability~\cite{dharmakeerthi2014impact, calearo2019grid}, improve energy management ~\cite{huang2024unveiling} and dynamic pricing strategies~\cite{kazemtarghi2024dynamic, bernard2025dynamic}, and shape policies~\cite{mastoi2022depth, li2016electric}, allowing a reliable and cost-effective transition to electric mobility. Forecasting charging demand is therefore not only a matter of operational efficiency, but also a critical lever for accelerating decarbonization.

However, despite growing interest~\cite{yousuf2024depth}, three persistent challenges remain. First, most prior work relies on small-scale datasets that do not capture the intricate complexity of real-world charging patterns across locations and usage scenarios~\cite{van2024large, martins2022short, huttel2021deep}. Second, existing models often treat charging sites in isolation~\cite{van2024large}, forecast regional demand ~\cite{louie2017time, yi2022electric, wang2018application, kim2021forecasting}, or cluster by proximity~\cite{miltner2024towards}, unable to model complex temporal and spatial relationships between sites. Third, there is limited understanding of how global knowledge can be transferred to forecast demand at newly deployed sites where there is little to no operational history~\cite{rashid2024comprehensive}. Inaccurate demand modeling not only inflates operational costs, but also slows EV adoption---challenges that can be mitigated through informed interventions such as curtailment, peak-shaving through dynamic pricing, and optimized battery dispatch.

This work leverages a novel industry-level dataset covering most U.S. public Level 3 (DC fast) charging sites, enabling a fine-grained analysis of market behavior and temporal patterns previously unattainable. Building on this foundation, three key contributions are made:
(1) Site archetypes are identified by combining clustering with few-shot forecasting, with the number of clusters selected based on predictive performance at unseen sites.
(2) Semantic characterizations of the archetypes using a nationwide dataset, linking clusters to geography, surrounding amenities, and usage contexts.
(3) Demonstrating knowledge transfer to new sites with limited history, showing that archetype-specific models consistently outperform global models in few-shot forecasting.

\begin{figure}
    \centering
    \includegraphics[width=1.0\linewidth]{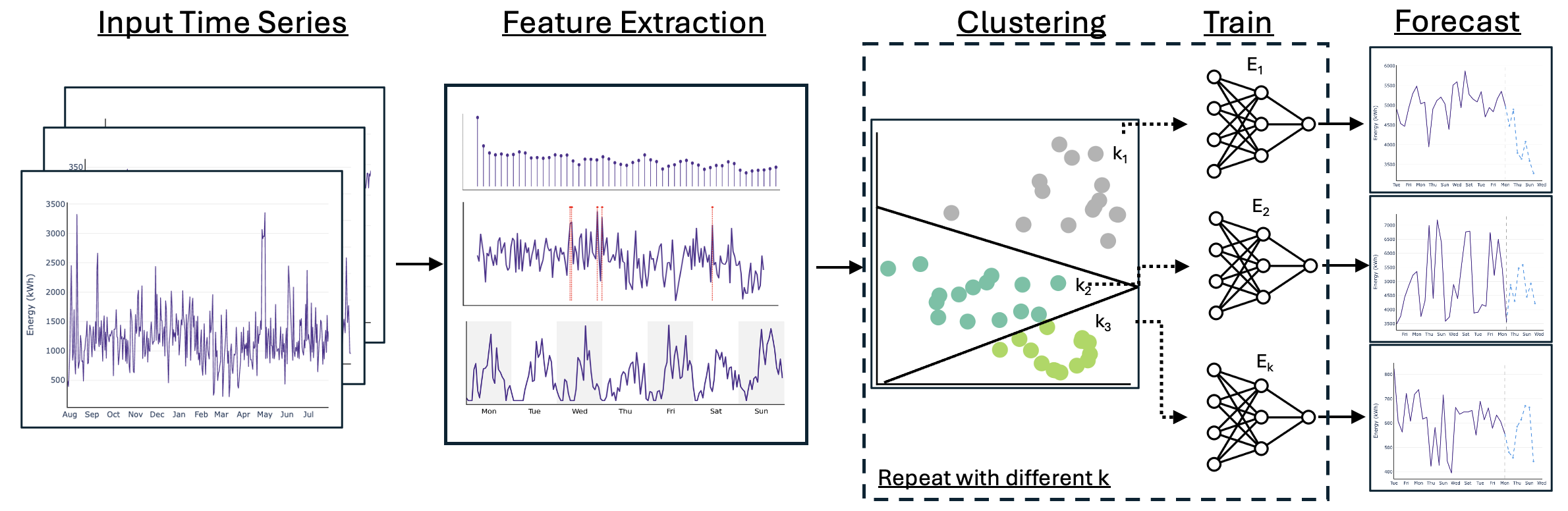}
    \caption{The framework leverages historical demand data to generate features for clustering, with expert models trained on resulting site clusters to forecast demand. This is repeated for different $k$. }
    \label{fig:methodology}
\end{figure}

\section{Methodology}
Consider a dataset represented as $X=\{x_1, x_2,\dots,x_n\}$,
where each time series $x_i=\{(t_{ij}, y_{ij})\}^{T_i}_{j=1}$ corresponds to the site $i$, $t_{ij}$ denotes the calendar date of observation $j$, and $y_{ij}$ is the total energy sold (in kWh) at the site $i$ on day $t_{ij}$. The dataset is partitioned into training and test sets, $X_{train}$ and $X_{test}$, using a standard 80\%--20\% split. For $X_{test}$, the historical context is restricted to $L=28$ days (four weeks) preceding the forecast horizon. For $X_{train}$, the complete history per site is used (ranging from several weeks to multiple years, depending on availability). 
The objective is to learn a predictive mapping such that, for each unseen site, the model (or models) produces a forecast:
\begin{equation}
\hat{\mathbf y}_{t+1:t+H \mid t} \;=\; f_\theta\!\left(\mathbf y_{t-L+1:t}\right),
\end{equation}
where $y_{t-L+1:t} = (y_{t-L+1},\dots,y_t)$ denotes the observed demand in the context window, and $f_{\theta}$ is the forecasting function parameterized by $\theta$. We set the forecast horizon $H=7$ days to capture both weekday and weekend demand while aligning with weekly cycles in energy management, dynamic pricing, and operational planning. This setup evaluates ``few-shot'' inference, as lag $L=28$ is statistically insignificant to capture seasonality and variability in demand. 

To achieve this, we characterize site-level demand by aggregating into weekly utilization profiles. In addition, we extract a set of canonical features (e.g., distributional characteristics, autocorrelation structure, and outlier dynamics), summarizing key statistical properties of each time series~\cite{Lubba2019:Catch22CAnonicalTimeseries}. Using these representations, we apply k-means~\cite{lloyd1982least} to group sites into clusters of similar behavior. 
For each candidate number of clusters $k \in \{1,\dots,20\}$, cluster assignments are used to train a Temporal Fusion Transformer~\cite{lim2021temporal} model specialized for that cluster, resulting in a set of forecasting models $\Theta=\{\theta_1,\dots,\theta_k\}$. The performance of the model is then evaluated across $k$ to identify the optimal clustering.
Symmetric Mean Absolute Percentage Error (sMAPE) and Root Mean Squared Error (RMSE) are used to capture complementary aspects of predictive performance, where sMAPE measures relative accuracy across scales, and RMSE emphasizes magnitude and sensitivity to large deviations. 
Note that similar performance gains are observed with other architectures such as TCN~\cite{bai2018empirical} and N-BEATS~\cite{oreshkin2019n} with varying baseline (or global) performance.

This experimental design enables us to evaluate:
(i) generalization to previously unseen sites,
(ii) the relative performance gains of cluster-specialized ``expert'' models compared to a global baseline, and
(iii) the appropriate number of clusters (archetypes) that best capture charging behavior. Figure~\ref{fig:methodology} shows an overview of the framework. 

\section{Key Results}

\begin{figure}
    \centering
    \includegraphics[width=0.85\textwidth]{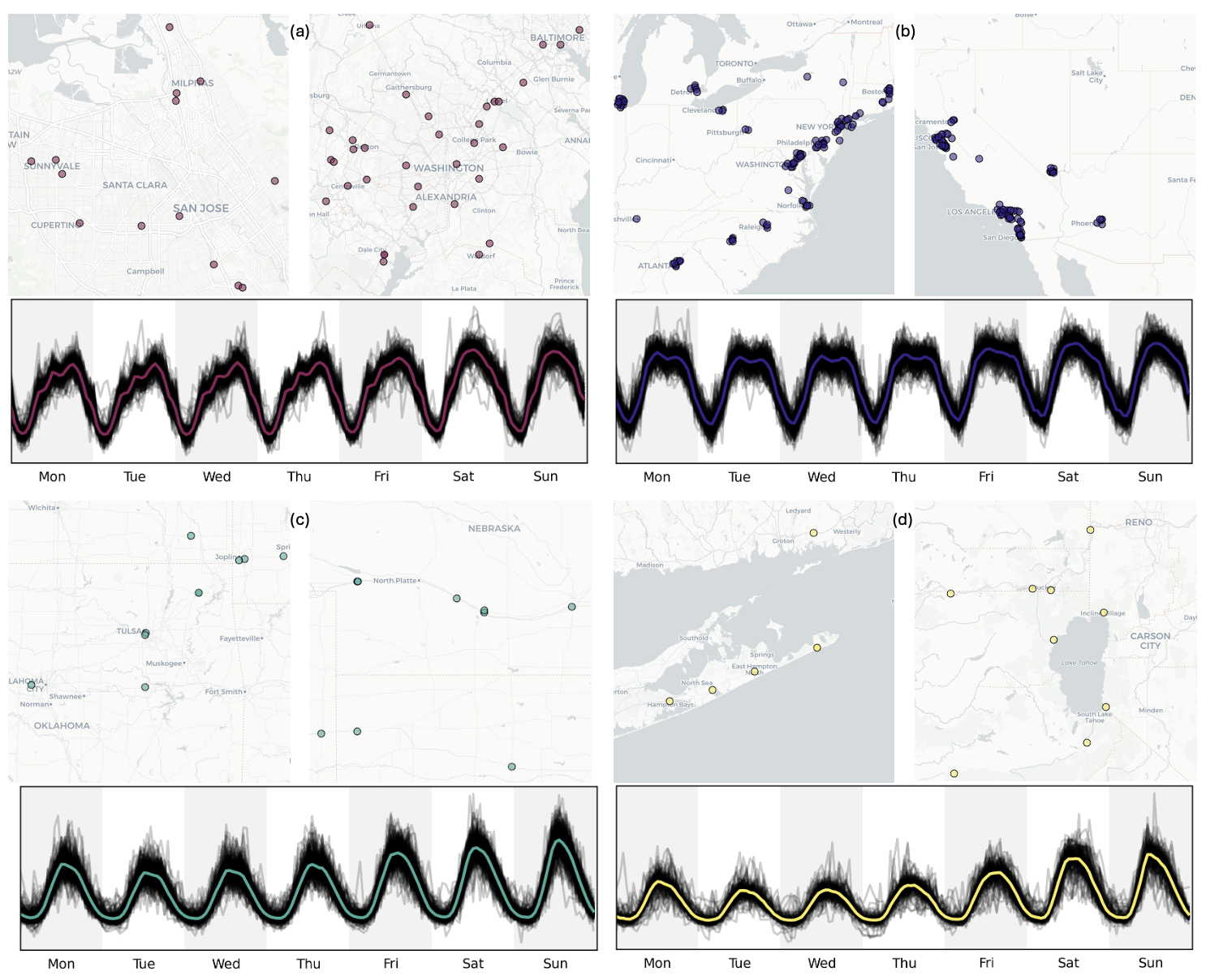}
    \caption{\textbf{(a) A8}: Recurring weekday patterns across downtown cores;  \textbf{(b) A1}: Stable daytime demand in metro areas adjacent to retail chains;  \textbf{(c) A3}: Utilization concentrated along major highways; \textbf{(d) A11}: Elevated weekend demand near leisure hubs such as Lake Tahoe and Hampton Bays. See Appendix~\ref{subsec:cluster_spread} for complete geospatial distribution and demand profiles. }
    \label{fig:geoplot}
\end{figure}
\textbf{Dataset:} The AGCharging dataset contains session data from 8000 DC Fast Charging (Level 3) sites filtered to the top five charging networks by utilization to capture dominant patterns. The charging sessions are aggregated at the daily level to produce a dataset containing each site's daily energy consumption. Sites with fewer than 35 days of history are excluded, as the model requires a 28-day lag and a 7-day prediction window. To support research, a slice of this dataset will be released.

\begin{wrapfigure}{r}{0.3\linewidth}
\centering
\includegraphics[width=\linewidth]{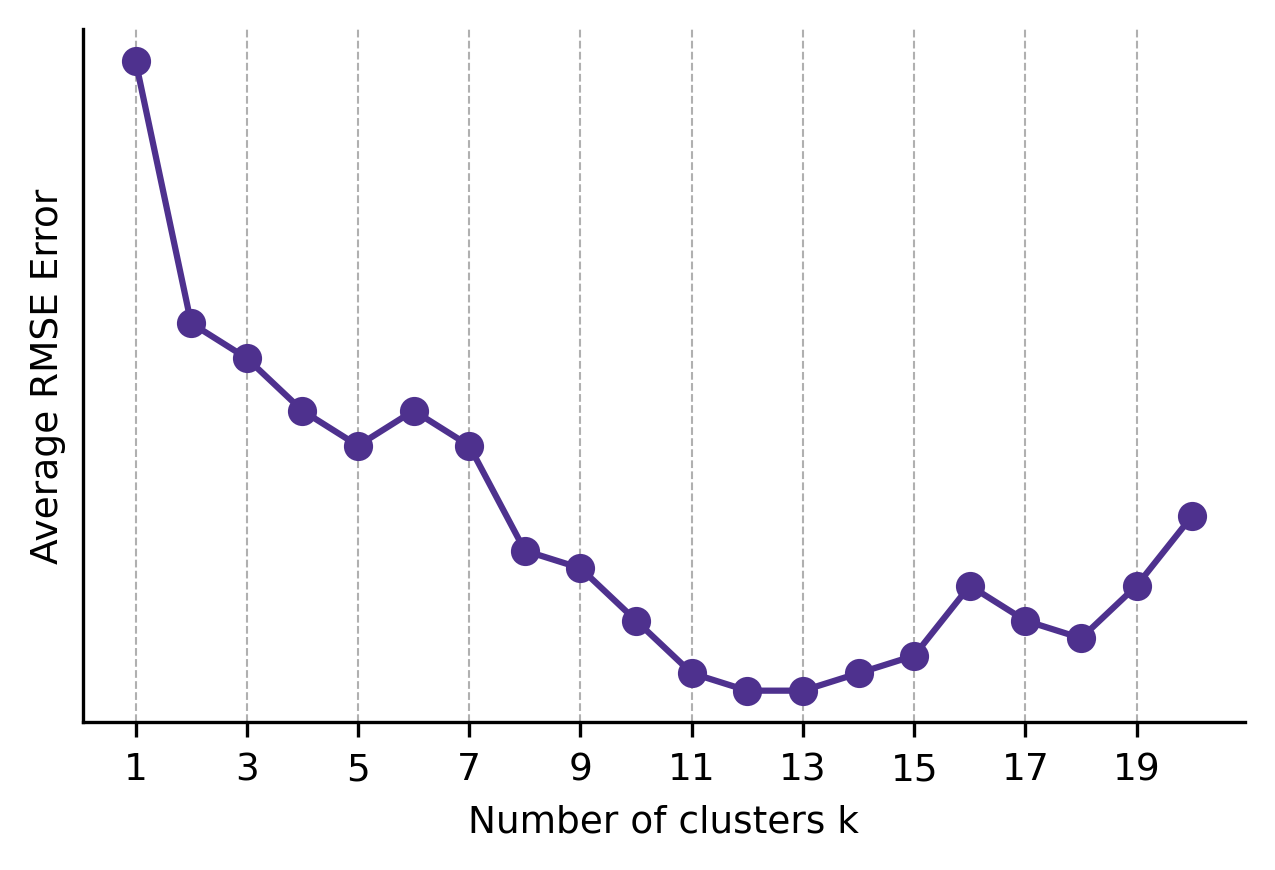}
\caption{Global $(k=1)$ vs expert $(k>1)$ performance.}
\vspace{-1.5em}
\label{fig:elbow}
\end{wrapfigure}

\textbf{Forecasting Performance:} We evaluate  accuracy on unseen sites in $X_{test}$ by first predicting cluster memberships using the k-means predictor. The expert models are then trained on different cluster counts $k$ to determine the optimal granularity.

Figure~\ref{fig:elbow} demonstrates the trade-off between global and expert models. The global model ($k=1$) provides a strong baseline, while specialized models yield substantial gains up to an optimal $k=12$. Beyond this point, excessive clustering reduces predictive power, highlighting the importance of balancing cluster size. This suggests that twelve distinct site archetypes capture dominant demand patterns and enable robust few-shot inference.

To interpret these clusters, we attach semantic labels and visualize them on a raster map. Figure~\ref{fig:geoplot} shows accurate clustering results with clear behavioral patterns: the downtown sites exhibit weekday ramps driven by commuters, the travel corridor sites cluster together with weekend spikes and high variance, the amenity-adjacent sites show consistent retail-linked usage, and the vacation destination sites display deep weekday troughs and pronounced weekend peaks.

\begin{table}[ht]
    \setlength{\tabcolsep}{4pt} 
  \centering
  \caption{Archetypes and performance of global (G) vs. expert (E) models. Expert models consistently outperform, especially in volatile clusters. $\sigma$ denotes deviation from baseline demand. }
  \label{main_table}
  \begin{tabular}{p{9.5cm}cc|cc}
    \toprule
    \multirow{2}{*}{Archetype (\% of sites) and Description} & \multicolumn{2}{c}{sMAPE} & \multicolumn{2}{c}{RMSE} \\
    \cmidrule(lr){2-3} \cmidrule(lr){4-5}
     & G & E & G & E \\
    \midrule
    \textbf{A1. Steady Retail (7\%):}  
    Stable all-week 9AM-5PM demand ($\uparrow 1.3\sigma$), predominantly located near retail chains in cities such as Los Angeles, Chicago, Denver, and Washington DC. 
     & 16.15 & \textbf{15.82} & \textbf{614} & 616 \\
    
    \textbf{A2. Urban Corridors (12\%):}
    Large weekend peaks ($\uparrow 2.6\sigma$) co-located with travel plazas near popular destinations and metros. 
     & 24.17 & \textbf{20.76} & 728 & \textbf{654} \\

    \textbf{A3. Regional Corridors (12\%):}  
    Similar peaks ($\uparrow 2.1\sigma$) to A2 but lower baseline and higher variance, reflecting sparser corridors. 
     & 30.23 & \textbf{28.60} & 638 & \textbf{608} \\

    \textbf{A4. Mixed Urban (12\%):}  
    Blend of commuter and leisure usage ($\uparrow 1.5\sigma)$, observed in dense downtowns with diverse land use. 
     & 24.77 & \textbf{22.87} & 799 & \textbf{735} \\

    \textbf{A5. Balanced Urban (8\%):}
    Similar to A1, but cleaner day/night split, moderate peaks ($\uparrow1.21\sigma$), and deep troughs ($\downarrow1.75\sigma$). 
     & 23.46 & \textbf{21.80} & 650 & \textbf{608} \\

    \textbf{A6. Commuter Corridors (15\%):}  
    Weekday AM/PM ramps exhibit smoother peaks ($\uparrow1.40\sigma$) compared to sharper weekend peaks ($\uparrow1.9\sigma$), reflecting work–home travel rhythms.
     & 24.50 & \textbf{23.23} & 614 & \textbf{575} \\

    \textbf{A7. Mega Metro (4\%):}  
    Consistent, saturated ($\uparrow 1.0\sigma)$ profile characteristic of large EV markets such as Southern California, reflecting large-scale adoption. 
     & \textbf{12.30} & 12.91 & 1126 & \textbf{1005} \\

    \textbf{A8. Weekday Ramps (7\%):}  
    Pronounced weekday ramps with evening plateaus ($\uparrow 1.5\sigma$), seen in  San Francisco, Dallas and Seattle.
     & 12.72 & \textbf{8.69} & 1232 & \textbf{1131} \\

    \textbf{A9. Suburban Shopping (13\%):} 
    Spread across most states with midday peaking ($\uparrow1.55\sigma$), tied to grocery/Big Box stores. 
     & 17.86 & \textbf{17.19} & 829 & \textbf{792} \\

    \textbf{A10. Emerging Metro (3.5\%):}  
    Found in Houston, Atlanta, Orlando, and Dallas, where workplace, retail, and leisure usage combine but baseline demand remains volatile.($\downarrow 1.76\sigma$ and $\uparrow 1.21\sigma$). 
     & 33.67 & \textbf{32.89} & 682 & \textbf{664} \\

    \textbf{A11. Seasonal Leisure (4\%):}  
    Stronger weekend/holiday surges ($\uparrow 2.3\sigma$) than weekdays ($\uparrow1.3\sigma$), near resorts and vacation homes. 
     & 34.48 & \textbf{31.73} & 762 & \textbf{705} \\

    \textbf{A12. Erratic (2.5\%):}  
    Low baseline demand with irregular peaks ($\uparrow1.24\sigma$) and troughs ($\downarrow0.94\sigma$).  
     & 65.46 & \textbf{61.16} & 457 & \textbf{377} \\
    \bottomrule
  \end{tabular}
\end{table}

Table~\ref{main_table} summarizes the characteristics of each archetype along with comparative forecast performance. The results show that expert models consistently outperform the global model, confirming that local specialization improves predictive accuracy. The improvement is especially pronounced for highly variable sites, such as travel corridors and vacation destinations, where shared temporal patterns (e.g., holiday and event-driven spikes) allow the expert models to generalize effectively.

Overall, these results provide strong evidence that site-level archetypes are not only semantically interpretable, but also predictively useful, demonstrating the value of cluster-aware modeling for scalable forecasting of EV charging demand. See Appendix~\ref{subsec:cluster_profiles} for detailed profiles.

\section{Conclusion and Future Work}
\label{conclusion}

We present the first nationwide assessment of the U.S. public fast-charging market, highlighting utilization patterns and site-level heterogeneity. Using forecast-guided evaluation, we identify the optimal number of archetypes and show that cluster-aware expert models outperform global models in few-shot inference. Beyond accuracy gains, these insights strengthen interconnection studies, guide incentives toward underserved areas, and inform underwriting through archetype membership. Future work will extend to longer-term horizons, soft clustering, and  external signals (e.g., events). These directions will further enhance our ability to anticipate and manage the rapidly evolving EV charging ecosystem.

\small
\bibliographystyle{unsrtnat}
\bibliography{references}

\newpage
\appendix
\section{Appendix}
\label{sec:appendix}
\subsection{Cluster spread across regions}
\label{subsec:cluster_spread}
\begin{figure}[h]
    \centering
    \includegraphics[width=1.0\textwidth]{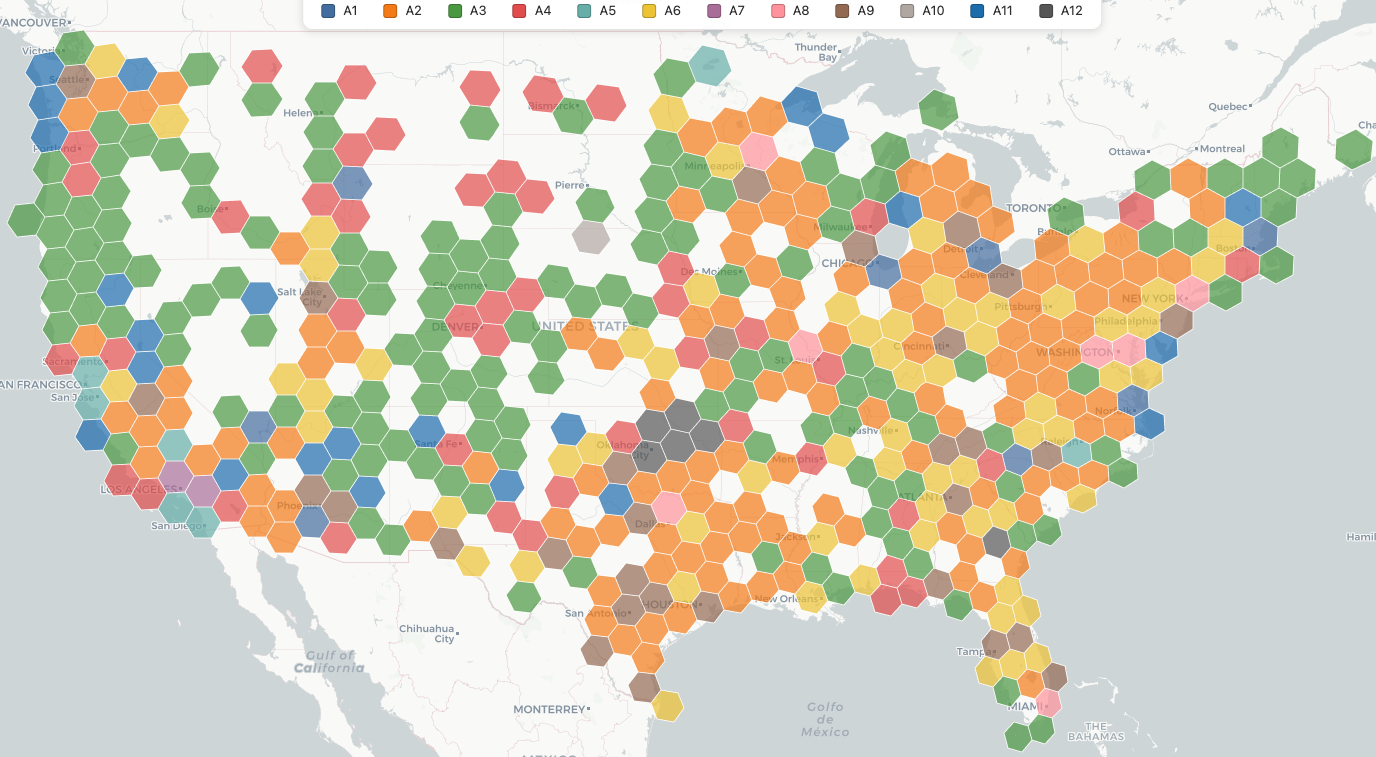}
    \caption{Geospatial distribution of charging site archetypes across the United States. Sites are aggregated into hexagonal cells, with each hexagon colored by the dominant utilization cluster.}
    \label{fig:hexmap}
\end{figure}

Figure~\ref{fig:hexmap} visualizes the geospatial distribution of charging site archetypes across the US. Individual sites are aggregated into hexagonal cells~\cite{uber_h3_docs}, and each hexagonal is colored by the dominant utilization cluster within its boundary. The result highlights regional demand patterns, from dense urban corridors to steady retail hubs and emerging suburban clusters. By abstracting sites into cluster-level patterns, the map provides a high-level view of infrastructure heterogeneity that informs siting, forecasting, and pricing strategies.

\subsection{Cluster Demand Profiles}
\label{subsec:cluster_profiles}
\begin{figure}[h]
    \centering
    \includegraphics[width=1\textwidth]{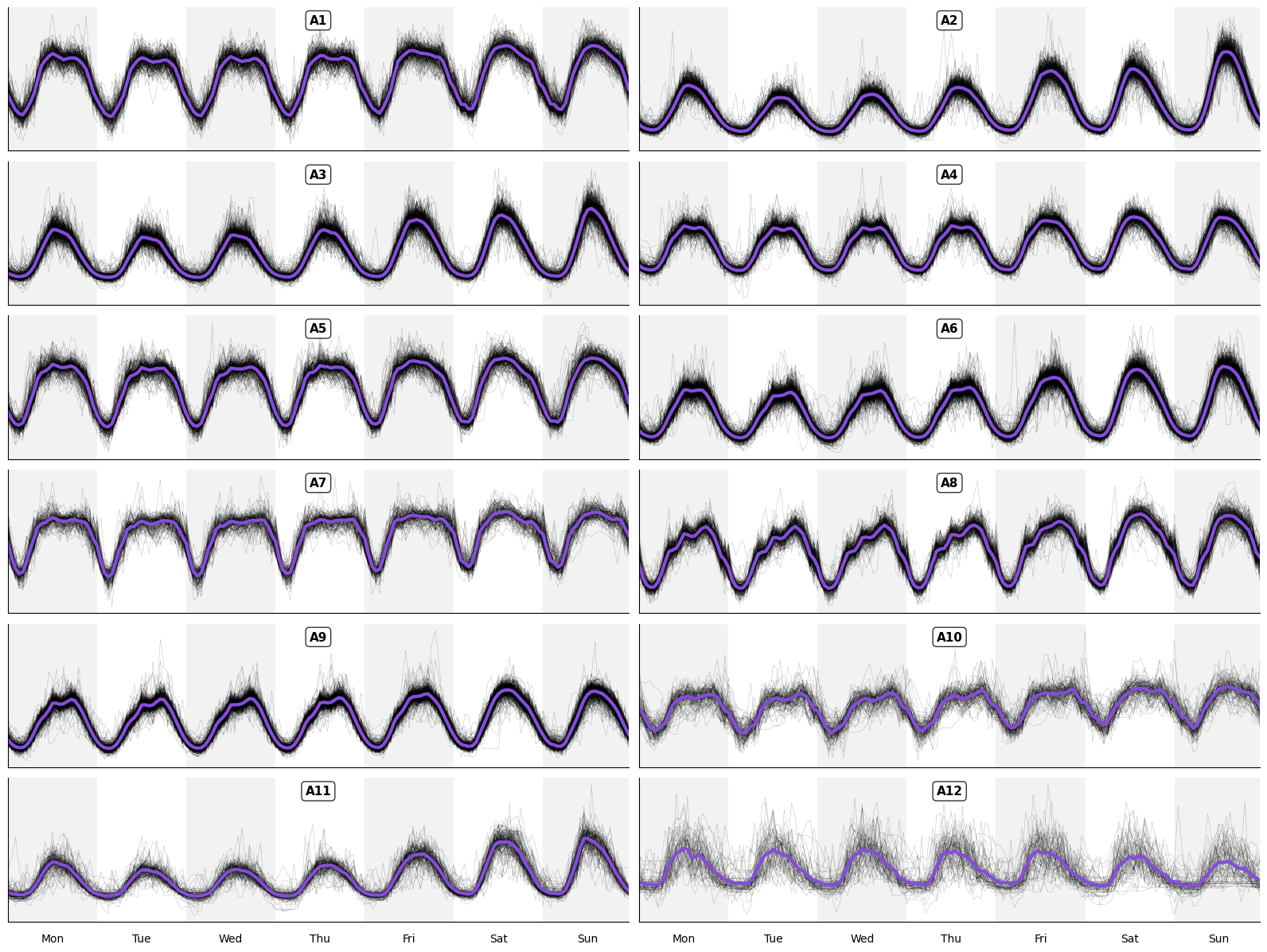}
    \caption{Representative demand profiles for the  identified archetypes. Each profile shows the barycenter of clustered sites, illustrating characteristic temporal patterns ranging from stable weekday cycles to highly volatile weekend or tourism-driven surges.}
    \label{fig:cluster_profiles}
\end{figure}

Figure~\ref{fig:cluster_profiles} highlights 12 distinct demand profiles that capture the diversity of EV charging behaviors throughout the US. Patterns range from regular weekday commuter flows (e.g., A8 and A10) to volatile and event-driven surges (A11). Stable archetypes such as A1 and A5 exhibit smooth and predictable demand, while clusters like A2, A3, A4, and A6, blend variability from both commuter and leisure segments. At the extremes, A7 reveal unique large-market dynamics, while A9 illustrates widespread midday–oriented demand tied to retail activity. Together, these profiles provide a structured lens on how location type and land-use context shape charging demand.

\subsection{Forecasting Performance}
\begin{figure}
    \centering
    \includegraphics[width=1\textwidth]{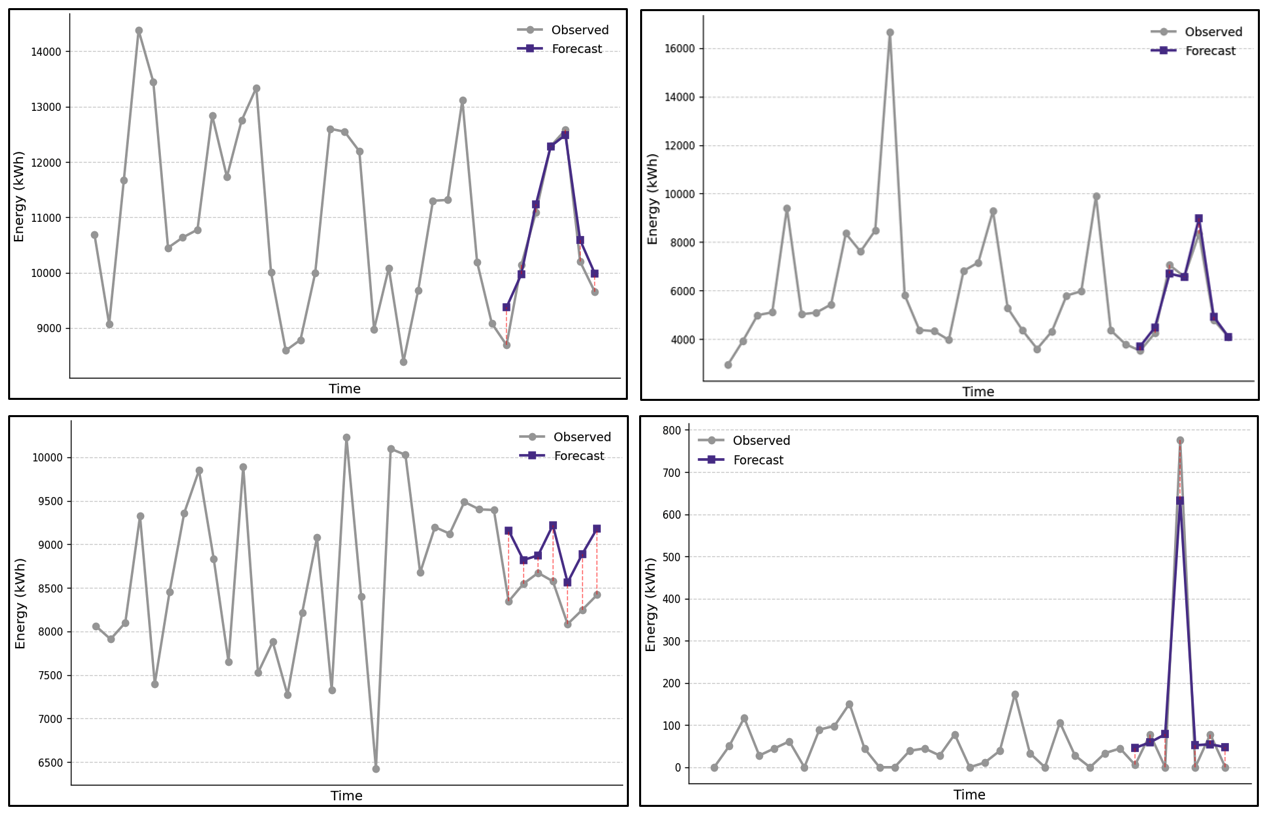}
    \caption{Examples of model forecasts across diverse site profiles.}
    \label{fig:forecasting}
\end{figure}
The model demonstrates strong generalization across sites with distinct temporal patterns. As seen in Figure~\ref{fig:forecasting}, the model accurately identifies the timing and shape of peaks, though occasional deviations in magnitude remain. In some cases, the model predicts the correct demand shape while slightly deviating in scale. In particular, the model successfully infers peaks in settings where no such peaks were observed in the historical data, highlighting its transferability and robustness.

\end{document}